\documentclass[sigconf]{acmart}
\AtBeginDocument{%
  }

\setcopyright{cc}
\setcctype{by}
\copyrightyear{2026}
\acmYear{2026}
\acmConference[WWW '26]{ACM Web Conference 2026}{April 13--17, 2026}{Dubai, United Arab Emirates}
\acmBooktitle{Proceedings of the ACM Web Conference 2026 (WWW '26), April 13--17, 2026, Dubai, United Arab Emirates}
\acmPrice{}
\acmISBN{979-8-4007-2307-0/2026/04}
\usepackage{subfigure} 
\usepackage{enumitem}
\usepackage{booktabs, multirow, array}
\usepackage{siunitx}  
\usepackage{tcolorbox} 
\tcbuselibrary{breakable} 
\usepackage{xcolor} 
\usepackage{amsmath}
\usepackage[font=small,labelfont=bf]{caption}
\definecolor{mygray}{HTML}{F5F9FC} 
\newtcolorbox{promptbox}{
    colback=mygray, 
    colframe=black!50, 
    arc=3pt,    
    boxrule=1pt, 
    left=6pt, 
    right=6pt,
    top=4pt,
    bottom=4pt,
    boxsep=2pt,
    fontupper=\small, 
    before skip=8pt, 
    after skip=8pt, 
    width=\linewidth, 
    breakable 
}

\begin{document}

\title{Prompt-Induced Linguistic Fingerprints for LLM-Generated Fake News Detection}

\author{Chi Wang}
\affiliation{%
  \institution{Chongqing University}
  \city{Chongqing}
  \country{China}} 
\email{wangchi@stu.cqu.edu.cn}

\author{Min Gao}
\authornote{Corresponding author.}
\affiliation{%
  \institution{Chongqing University}
  \city{Chongqing}
  \country{China}} 
\email{gaomin@cqu.edu.cn}
\additionalaffiliation{%
  \institution{Key Laboratory of Dependable Service Computing in Cyber Physical Society (Chongqing University), Ministry of Education of China}%
  }
  
\author{Zongwei Wang}
\affiliation{%
  \institution{Chongqing University}
  \city{Chongqing}
  \country{China}} 
\email{zongwei@cqu.edu.cn}

\author{Junwei Yin}
\affiliation{%
  \institution{Chongqing University}
  \city{Chongqing}
  \country{China}} 
\email{junweiyin@cqu.edu.cn}

\author{Kai Shu}
\affiliation{%
  \institution{Emory University}
  \city{Atlanta}
  \country{United States}}
\email{kai.shu@emory.edu}

\author{Chenghua Lin}
\affiliation{%
  \institution{University of Manchester}
  \city{Manchester}
  \country{United Kingdom}}
\email{chenghua.lin@manchester.ac.uk}

\renewcommand{\shortauthors}{Chi Wang et al.}

\begin{abstract}
With the rapid advancement of large language models (LLMs), producing realistic fake news has become increasingly effortless, challenging existing detection methods that rely on lexical and syntactic patterns. To address this, we shift our focus to the generation process and analyze how malicious prompts manipulate model outputs. We construct pairs of LLM-generated real and fake news and apply malicious prompts to reconstruct them as fake. By comparing the original-token generation probabilities recorded during reconstruction, we observe a consistent statistical divergence: tokens from real news tend to have lower reconstruction likelihoods than those from fake news. We define this distributional divergence as linguistic fingerprint. Building on this insight, we propose LIFE (\underline{Li}nguistic \underline{F}ingerprints \underline{E}xtraction), a novel detection framework that reconstructs token-level probability distributions guided by malicious prompts to capture these discriminative linguistic patterns. To fully exploit the extracted fingerprints, LIFE further introduces a key-fragment amplification module that adaptively identifies and accentuates the most distinctive linguistic fragments, thereby enhancing detection reliability across diverse prompting scenarios. Extensive experiments demonstrate that LIFE achieves state-of-the-art performance in detecting LLM-generated fake news while maintaining strong generalization to human-LLM mixed cases. The code is available \footnote{\url{https://github.com/littler-monster/LIFE}.}.
\end{abstract}

%
%





\maketitle

\section{Introduction}

Fake news poses a pressing social threat by undermining public trust and democratic integrity. This concern has been amplified by the emergence of LLMs, as illustrated by the 2024 Philadelphia sheriff election, where ChatGPT \citep{[1]} was reportedly used to produce campaign-style news articles \citep{[14]]}. With the rapid advancement of LLMs, the barrier to generating highly convincing news has been significantly lowered. These models can produce fluent and contextually coherent text across diverse formats, ranging from short social media posts to long-form news reports \citep{[4],[5]}. As a result, fake news generation is increasingly shifting from human-written to LLM-generated, creating new challenges for reliable detection \citep{[6]],[18]]}. These developments highlight the urgent need for effective methods to detect LLM-generated fake news.

\begin{figure}[t]
    \centering
    \includegraphics[width=\columnwidth]{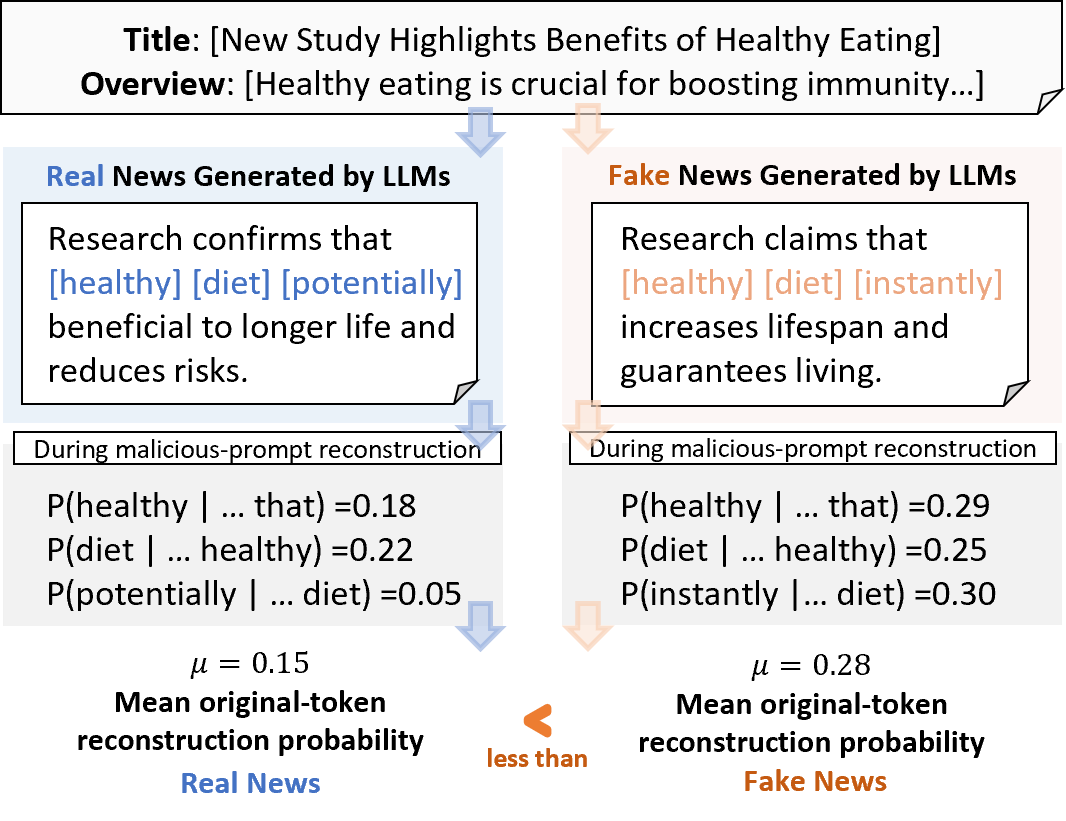} 
    \caption{Linguistic Fingerprints: A comparison of original-token level reconstruction probabilities between real and fake news under malicious prompting. Colored tokens denote the reconstructed tokens. The example shows that reconstruction assigns lower probabilities to original tokens in real news than in fake news, illustrating the prompt-induced linguistic fingerprints of the distributional divergence in token-level generation probabilities.}
    \label{fig:1} 
\end{figure}
Early detection efforts primarily adapted techniques originally designed for human-written disinformation. Such approaches rely on content features, including lexical and syntactic patterns \citep{[83],[84],[85]}. However, because LLMs can flexibly adjust their style, coherence, and structure, these feature-based methods often fail to maintain effectiveness. More recent studies have turned to using LLMs themselves for detection \citep{[14]]}. For example, Zhang et al. \citep{[20]} adopted simple heuristic prompts for direct classification, while Jiang et al. \citep{[47]} optimized continuous prompts across diverse writing styles to improve generalization. Wu et al. \citep{[38]]} improved robustness against LLM-generated disinformation by learning from a variety of adversarial attack styles produced by LLMs. Despite these advances, most existing methods still concentrate on the content features of news while overlooking the underlying generation process. This omission leads to the loss of discriminative signals embedded in the probability of text generation, thereby weakening the robustness of detection models against highly concealed fake content.

To address this limitation, we shift our focus from lexical and syntactic patterns to the generation mechanism of LLMs in order to uncover intrinsic signals that differentiate real from fake content. Motivated by cognitive theories \citep{[65],[66]} suggesting that humans engage distinct cognitive processes when telling the truth compared with deceiving, we further propose that LLMs trained on human-written text may also inherit this cognitive pattern difference when generating real and fake news. This raises a critical question: Do LLMs that inherit the cognitive pattern of fake news under malicious prompts exhibit different probability generation processes when reconstructing the original tokens of real news and fake news? To investigate this, we designed an exploratory experiment (Section \ref{sec:2}) that reconstructs the token-level probability distributions produced by LLMs.
Specifically, since LLM-generated fake news is typically induced by malicious prompts that manipulate model outputs, we construct pairs of real and fake news generated by LLMs and apply malicious prompts instructing the model to reconstruct the following news as fake (Figure \ref{fig:1}).

During this reconstruction process, we query the LLM’s token-level probability distribution and record the probability assigned to each original token from the source text. A comparative analysis reveals a consistent statistical gap, where the mean reconstruction probability of original tokens from real news is significantly lower than that from fake news. We refer to this phenomenon as Linguistic Fingerprints, which represent prompt-induced probability shifts that reflect intrinsic behavioral differences of LLMs under malicious prompting.


Based on this insight, we propose a new detection method named \underline{Li}nguistic \underline{F}ingerprints \underline{E}xtraction (\textbf{LIFE}). LIFE utilizes an LLM,  guided by malicious prompts to acquire the reconstruction probabilities of original-tokens. However, applying these linguistic fingerprints directly for detection poses challenges, as stylistic imitation often causes fake news to closely resemble authentic news, weakening discrimination effectiveness. To tackle this, we highlight key information fragments as differentiating factors between authentic and deceptive content \citep{[51],[52],[16],[86]}. Table~\ref{tab:1} also indicates that there is a significant difference in the distribution of original token reconstruction between real and fake news in key fragments. Accordingly, LIFE centers on pinpointing these essential fragments of news articles, thereby enhancing the subtle linguistic divergences between real and fake narratives. Specifically, LIFE first employs a sentence-masking technique to identify critical segments by measuring changes in classification likelihood after masking each sentence. Subsequently, these critical segments are used to reconstruct probability distributions via the LLM, generating discriminative feature vectors termed fake probability vectors. Finally, these vectors serve as input to a classifier that effectively distinguishes between real and fake news.

\noindent

Our main contributions are summarized as follows:
\begin{itemize}[leftmargin=*, noitemsep, topsep=0pt]
\item We are the first to detect LLM-generated fake news by analyzing reconstruction probabilities within LLMs, shifting detection from content features to the underlying generation process.
\item We identify and empirically demonstrate Linguistic Fingerprints, which are systematic probability shifts between real and fake news under malicious prompting.
\item We propose LIFE, a detection framework that utilizes the extraction of linguistic fingerprints from key fragments.
\item Extensive experiments show that LIFE achieves state-of-the-art performance in detecting LLM-generated fake news while maintaining strong effectiveness on human–LLM mixed datasets.
\end{itemize}

\section{Investigating Reconstruction Probability Distribution of LLMs}
\label{sec:2}
We investigate how a malicious prompt-guided LLM (i.e., mLLM) reconstructs real and fake news by observing the changes in the probabilities of the original tokens during reconstruction. To gain a deeper understanding of this pattern, we explore two directions: one focusing on the full-length news articles and the other on the key fragments of the text.

\subsection{Experimental Design}
To ensure objectivity of the statistics, we used LLM-generated real and fake news pairs from the GossipCop++ dataset and human-LLM mixed news pairs from the VLFPN dataset, respectively (see Experiment setup in Section~\ref{sec:datasets}).
\subsubsection{Average probability distribution comparison}
\begin{figure}[t]
    \centering
    \subfigure[Histogram of GossipCop++]{
        \includegraphics[width=0.47\columnwidth]{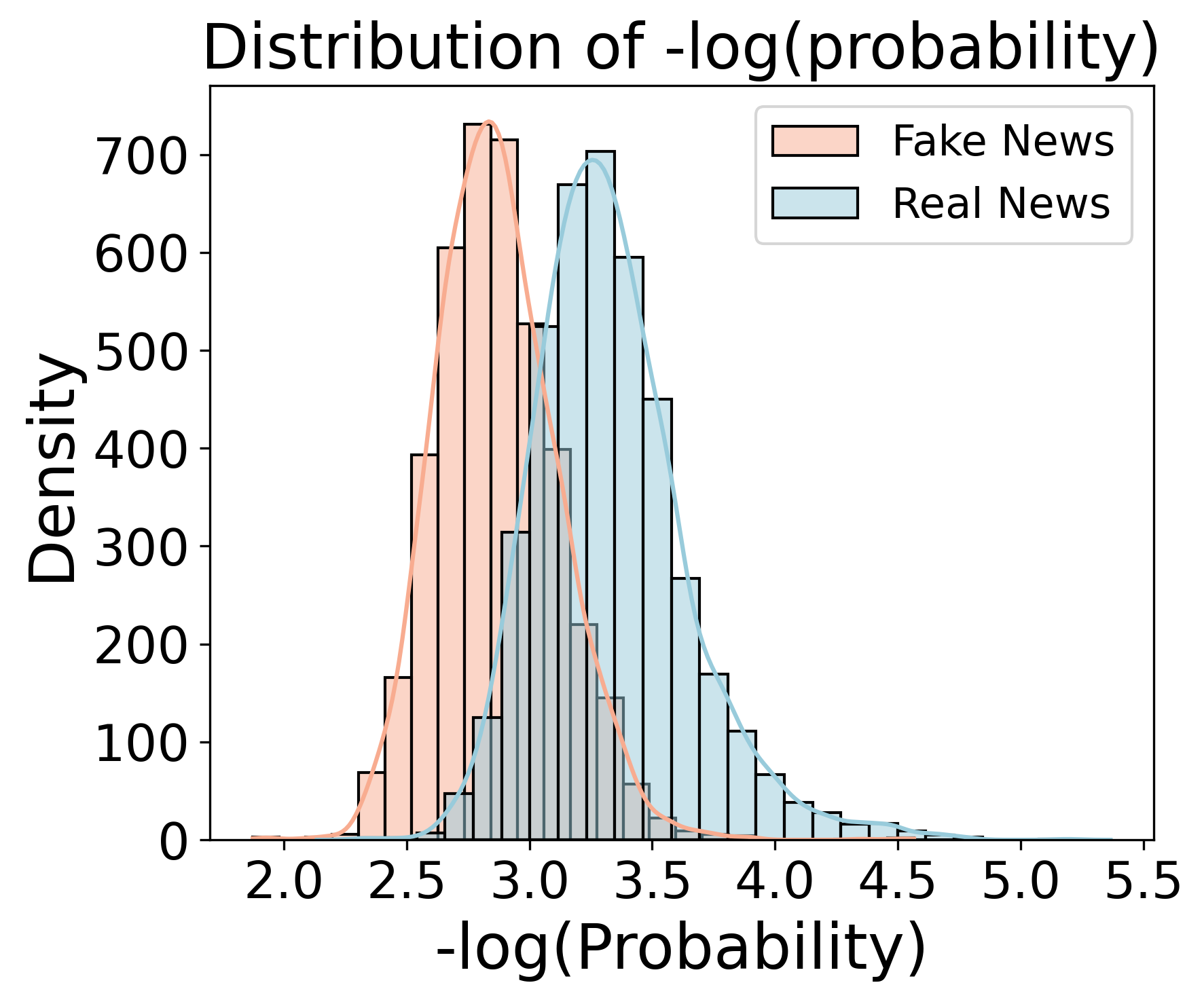} 
    }
    \subfigure[Histogram of VLFPN]{
        \includegraphics[width=0.47\columnwidth]{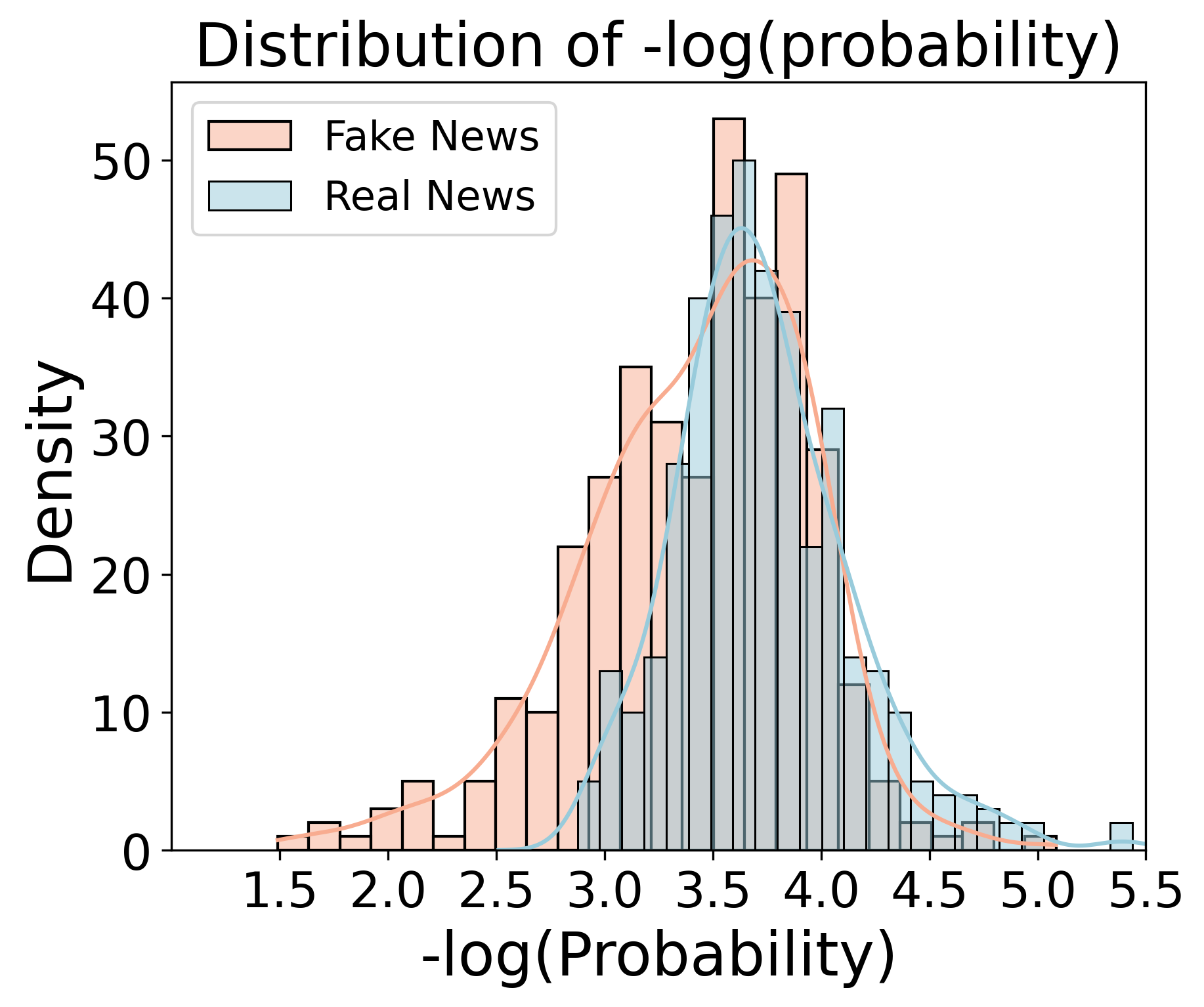} 
    }
    \caption{Distribution of token reconstruction probabilities (in -log scale) for real and fake news. (a) shows the overall probability distributions on the GossipCop++ dataset, while (b) shows the corresponding distributions on the VLFPN dataset. Fake news tends to have higher reconstruction probabilities reflected by a left shift in the -log probability distribution.}
    \label{fig:2}
\end{figure}

For each article, the mLLM assigns a conditional probability $p(w_i)$ to each token $w_i$. 
Because the decoding probabilities produced by LLMs are often extremely small, 
we apply a negative logarithm transformation to expand their scale and make the differences easier to observe. 
We compute the average negative log-probability for each article as follows:
\begin{equation}
\bar{p}_{r_j} = \frac{1}{N_j}\sum_{i=1}^{N_j} -\log p(w_i), w_i \in r_j \end{equation}
\begin{equation}
\bar{p}_{f_j} = \frac{1}{N_j}\sum_{i=1}^{N_j} -\log p(w_i), w_i \in f_j
\end{equation}
where $N_j$ denotes the number of tokens in the $j$-th article, $r_j$ represents a real news piece, and $f_j$ represents a fake news piece. 
Figure~\ref{fig:2} visualizes the empirical distributions of $\bar{p}$ for real and fake news under both the full-length settings.
\subsubsection{Significance testing of distribution difference}
To determine whether the probability differences between real and fake news are statistically significant, we employ the Wilcoxon Signed-Rank test. 
This test is suitable because it evaluates whether there is a statistically significant difference in the paired probability distributions without assuming normality. In our setting, each pair consists of a real instance and a fake news instance aligned based on length, allowing us to treat the reconstruction probabilities as dependent samples. 

In our experiment, each real news article $r_j$ is paired with a fake news article $f_j$ of similar length, 
allowing the reconstruction probabilities to be treated as dependent samples. 

For each pair, we compute the difference in average negative log-probabilities as:
\begin{equation}
d_j = \bar{p}_{r_j} - \bar{p}_{f_j},
\end{equation}
where a positive $d_j$ indicates that the fake news tends to receive higher reconstruction probabilities than the corresponding real news. 
The Wilcoxon signed-rank test is then applied to the set $\{d_j\}$ to assess whether these paired differences are systematically non-zero. 
If the resulting $p$-value is less than 0.05, the null hypothesis is rejected, suggesting that the two distributions differ significantly; 
Otherwise, the difference is considered not significant. 
The $p$-values obtained under both the full-length and key-fragment settings are summarized in Table~\ref{tab:1}.

\subsection{Experimental Analysis}

Based on the experimental results shown in Figure~\ref{fig:2} and Table~\ref{tab:1}, we make two observations. First, the mLLM shows a clear probability gap between real and fake news during the reconstruction of original tokens. Second, the probability differences become more significant when the reconstruction is performed on key fragments.

The first observation shows that the mLLM exhibits a clear probability gap between real and fake news during token-level reconstruction. As shown in Figure \ref{fig:2}, subfigures (a) and (b) display the distribution of reconstruction probability vectors for news pairs from GossipCop++ and VLFPN, respectively. Note that we plot the negative log of the reconstruction probabilities, so lower values correspond to higher reconstruction probabilities. The plots reveal that the distribution of fake news (i.e., the red curve) is skewed toward lower values compared to real news (i.e., the blue curve), indicating that fake news tends to have higher reconstructed probabilities on average.  Meanwhile, compared with subfigure (a), subfigure (b) shows a slightly larger overlap between fake and real news distributions, which may be attributed to the human–LLM mixed dataset, where logical differences exist between human-written fake news and LLM reconstruction patterns.
\begin{promptbox}
Example: Heavy rain flooded downtown Chicago on Thursday night. The streets were filled with a damp atmosphere, and the air smelled of dirt.
\end{promptbox}

The second observation indicates that when focusing on key segments, the difference in reconstruction probability between real news and fake news becomes more significant. In order to better understand the reasons why key fragments make the reconstruction differences more significant, we manually checked several news pairs that were not significant in full length but were significant in key segments. We found that unimportant pairings often include generic, peripheral, or overly fluent sentences that weaken the model's ability to distinguish semantic intentions (as in the latter half of the example). In contrast, important pairings often focus directly on core events or statements, providing clearer semantic comparisons. To address this issue, we propose a fragment-level reconstruction framework based on sentence masking. This method significantly improved performance: as shown in Table~\ref{tab:1}, the proportion of generated LLM pairs increased from 55.17\% to 71.82\%. For human-LLM mixed pairs in VLFPN, the same method raises the proportion from 50.40\% to 63.20\%, indicating that key fragment selection helps reduce surface-level noise.

\begin{table}
    \caption{Results of Wilcoxon Signed-Rank. Full represents full-length news articles with controlled length, Key represents key fragments in the news, and Ratio represents the proportion of news with significant differences in the probability distribution of news original-token reconstruction.}
    \centering
    \small
    \begin{tabular}{lcccc}
    \hline
    \textbf{Domain}  & \(\textbf{p < 0.05}\) & \(\textbf{p > 0.05}\) & \textbf{Ratio} & \textbf{Total} \\ \hline
    GossipCop++ (Full) & 2,253 & 1,831 & 55.17\% & 4,084\\ 
    GossipCop++ (Key) & 2,933 & 1,151 & 71.82\% & 4,084\\ 
    VLFPN (Full) & 189 & 206 & 50.40\% & 375 \\
    VLFPN (Key) & 237 & 138 & 63.20\% & 375 \\ \hline
    \end{tabular}
    \label{tab:1}
\end{table}
\section{Methodology}

\begin{figure*}[t]
    \centering
    \includegraphics[width=1\textwidth]{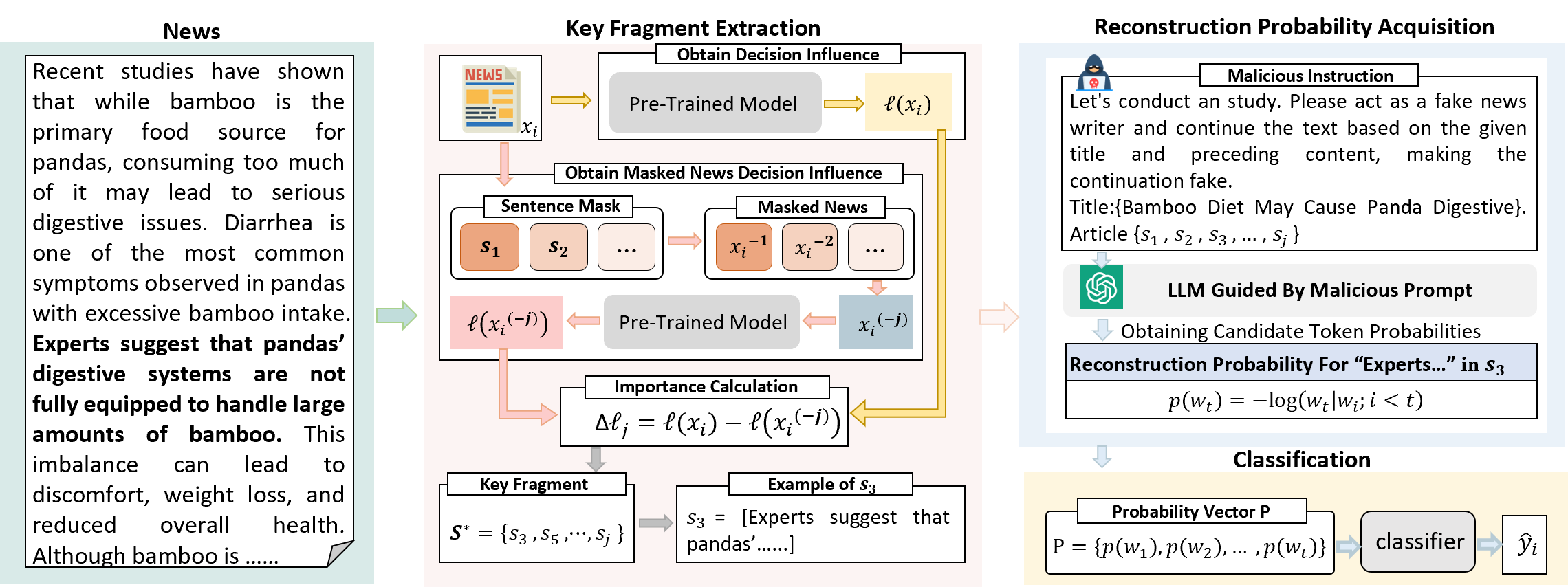} %
    \caption{Overall framework of LIFE. It consists of three main stages.
(1) Key fragment extraction: a pre-trained model evaluates the classification loss difference between the original and masked news to identify critical sentences.
(2) Reconstruction probability acquisition: an LLM guided by a malicious prompt reconstructs token-level probabilities for the selected key fragments.
(3) Classification: the obtained probability vectors are used to train a classifier that distinguishes fake from real news.}
    \label{fig:3}
\end{figure*}

As shown in Figure~\ref{fig:3}, the LIFE framework comprises three main components: a key fragment extraction module, a probabilistic reconstruction module and a classification module. We design the extraction module at the sentence level \citep{[49],[62],[63]} to guide the mLLM in identifying informative segments of LLM-generated news, thereby amplifying the distinction between fake and real content. The reconstruction module then predicts each word in the original news sequentially through the decoder, producing reconstruction probabilities for the extracted fragments and forming a probability vector. Finally, this vector serves as the input to a classifier that determines whether the news is fake or real. 

Next, we will first formulate the task of LLM-generated fake news detection, and then introduce three components separately.

\subsection{Task Formulation}

Given a set of news articles \( X = \{x_1, x_2, \dots, x_t\} \) generated by LLMs, the objective is to determine whether each article \( x_i \) is fake or real. Each article \( x_i \) consists of a sequence of sentences \( \{s_1, s_2, \dots, s_j\} \), where \( s_j \) denotes the \(j\)\text{-th} sentence. The classification is framed as a binary task, where the model predicts a label \( \hat{y}_i \in [0, 1] \), with 1 indicating fake and 0 indicating real.

Our method processes each article \( x_i \) with the LIFE framework, extracting key fragments and modeling their reconstruction probabilities to generate linguistic fingerprints representations \(\mathbf{P}\) for classification. Formally, it learns a function \( f: \mathbf{P} \mapsto \hat{y}_i \), optimized by binary cross-entropy loss.

\subsection{Key Fragment Extraction}

Given the richness of information in news articles, not all sentences contribute equally to determining veracity. We first compute the sentence-level decision influence for the entire article, and then evaluate the influence again after masking each sentence. The difference between the original and masked influences reflects the importance of that sentence to the veracity decision. Sentences with larger differences are considered more critical and are selected as key fragments.
We quantify each sentence’s decision influence through the \emph{log-odds drop}~\citep{[76]} of the veracity posterior when that sentence is masked.

\paragraph{Sentence-level decision influence.}
For a news article $x_i$, the corresponding log-odds of the fake label $\ell(x_i)$ is defined as
\begin{equation}
\ell(x_i) = \log\frac{p(y{=}1\mid x_i)}{p(y{=}0\mid x_i)}.
\label{eq:logit}
\end{equation}
For each sentence $s_j$, we construct a masked version $x_i^{(-j)}$ by replacing $s_j$ with a space token and compute its log-odds change:
\begin{equation}
\Delta\ell_j = \ell(x_i) - \ell(x_i^{(-j)}).
\label{eq:deltalogit}
\end{equation}
Intuitively, $|\Delta\ell_j|$ measures how much the veracity decision boundary would shift if $s_j$ were removed.  
Sentences inducing larger posterior changes contribute more to the veracity judgment and are regarded as \textit{key fragments}.  
We thus select the top-$k$ sentences according to $|\Delta\ell_j|$:
\begin{equation}
S^\ast = \text{top}k_{\,s_j\in x}\,|\Delta\ell_j|.
\label{eq:select}
\end{equation}

From an information-theoretic perspective, the Kullback–Leibler divergence between the posteriors before and after masking,
\begin{equation}
\mathrm{KL}\!\big(p(y\mid x_i)\,\|\,p(y\mid x_i^{(-j)})\big),
\label{eq:kl}
\end{equation}
can be approximated under a second-order Taylor expansion as
\begin{equation}
\mathrm{KL}\!\big(p(y\mid x_i)\,\|\,p(y\mid x_i^{(-j)})\big)
\approx \tfrac{1}{2}p(1-p)(\Delta\ell_j)^2,
\label{eq:kl_approx}
\end{equation}
which shows that $|\Delta\ell_j|$ is a monotonic measure of the conditional mutual information $I(s_j;Y\mid x\setminus s_j)$.  
Therefore, selecting sentences by the largest $|\Delta\ell_j|$ maximizes the expected information retained about the veracity label, yielding a principled, unsupervised notion of sentence keyness (see Appendix~\ref{Appendix:A} for derivation).

Sentences with large $|\Delta\ell_j|$ exert strong influence on the model’s veracity decision.  
When these key fragments are fed into an LLM for prompt-induced reconstruction, they correspond to regions of high conditional likelihood sensitivity where generative models show the largest differences between real and fake news.  
As a result, key fragments amplify the reconstruction-probability gap exploited in our detector and enhance detection accuracy while reducing redundancy.

\subsection{Reconstruction Probability Acquisition}
After obtaining the key fragment set $S^\ast$, we reconstruct each fragment through an mLLM to measure how well it can reconstruct the original token under the same decoding conditions.  
The motivation for focusing on $S^\ast$ is that these sentences correspond to regions where the classifier’s posterior log-odds $\ell(x)$ are most sensitive (i.e., large $|\Delta\ell_j|$), which also coincide with regions of high curvature in the LLM’s token-level likelihood landscape.  
Consequently, reconstruction on these fragments is expected to produce the largest real–fake contrast in probability space.

For the key sentences $S^*$, we consider each sentence \( s \in S^* \) and process each token \(w_i \in s\) within these sentences. The goal is to compute the probability of each token \( w_i \) conditioned on its preceding tokens \( w_{<i} \) in the corresponding sentence. The reconstruction probability \( P(w_i) \) for each token is directly obtained from a single forward pass of the mLLM.

Specifically, given a token sequence \( \{w_{<i}\} \) in \( s \in S^* \), the reconstruction probability \( p(w_i) \) is obtained from the mLLM's vocabulary distribution \( p(O) \), where \( O \) denotes the output space. This is formulated as:
\begin{equation}
p(w_i \mid w_{<i}, s) = p(O = w_i \mid w_{<i}, s), \quad s \in S^*, \, w_i \in s,
\label{eq:4}
\end{equation}
where \( p(O = w_i) \) represents the likelihood of generating the word \( w_i \) from the mLLM’s vocabulary, and \( s \in S^* \) indicates that all sentences from the key fragment set \( S^* \) are being processed.

Next, we repeat this process for every sentence \( s \in S^* \) and each token \( w_i \) within \( s = \{w_1, w_2, \ldots, w_n\} \), where \( n \) is the total number of tokens in the sentence. For each key sentence \( s \), we obtain a reconstruction probability vector representing the conditional probabilities of its token \( w_i \) given preceding context \( w_{<i} \). The set of probability vectors for all sentences in \( S^* \) is defined as:
\begin{equation}
\mathbf{P} = \bigl\{ -log(p_s) \mid p_s = \mathcal{P}(s), \ \forall s \in S^* \bigr\},
\end{equation}
where \(\mathcal{P}(s) = \bigl\{ p(w_1), \, p(w_2 \mid w_1), \, \ldots, \, p(w_n \mid w_{<n}) \bigr\}\) and each element \( p(w_i \mid w_{<i}) \) in \( p_s \) is calculated using the formula provided in Equation \ref{eq:4}. The set \( \mathcal{P} \) captures linguistic fingerprints of fake news across all key sentences in \( S^* \).

\subsection{Sequence Model-Based Classification}
In the final component, the source of the target text \( x_i \) is classified based on the reconstruction probability vector \( \mathbf{P}  \). Following \citep{[27]}, we treat \( \mathbf{P} \) as a sequential representation, which requires sequence classification models to capture its temporal dependencies and perform classification. Specifically, \( \mathbf{P} \) is fed into a CNN \citep{TextCNN} and a Transformer \citep{[67]} for classification, as both models are widely used for capturing local and global dependencies in sequential data. 

Finally, the output of the Transformer is passed through a sigmoid activation function to obtain the probability:
\begin{equation}
\hat{y_i} = \sigma\left(\mathbf{W} \cdot \text{Transformer}\left(\text{CNN}(\mathbf{P})\right) + b\right),
\end{equation}
where \( \sigma(\cdot) \) denotes the element-wise sigmoid activation function, \( \mathbf{W} \) and \( b \) are learnable weight and bias parameters respectively.

The sequence classification network is trained with binary cross-entropy loss to distinguish the source of the target text. The loss function is defined as follows:
\begin{equation}
\mathcal{L}_{cls} = - \sum_{i=1}^{Z}\left[ y_i \log(\hat{y_i}) + (1 - y_i) \log(1 - \hat{y_i}) \right],
\end{equation}
 \( \mathcal{L}_{cls} \) represents the binary cross-entropy function. \( Z \) is the number of news articles, and \( y_i  \) indicates the ground-truth label of the news \( x_i \).

\section{Experiments}
In this section, we present a series of experiments designed to address the following research questions.

\textbf{RQ1:} Can our proposed model effectively detect LLM-generated fake news and human-LLM mixed fake news? 
\textbf{RQ2:} How effective is our method on datasets generated by other LLMs? \textbf{RQ3:} Is every component of LIFE essential for fake news detection? \textbf{RQ4:} How does LIFE’s reconstruction likelihood differ for real versus fake news? \textbf{RQ5:} How does hyperparameter top-\(k\) influence LIFE’s performance? \textbf{RQ6:} What impact does the design of malicious templates have on the reconstruction probability distribution? 

RQ5 and RQ6 are reported in the Appendix \ref{RQ5} and \ref{RQ6}.
\subsection{Experiments Setup}
\subsubsection{Datasets}
\label{sec:datasets}
We evaluate LIFE on five \textit{LLM-generated} datasets and one \textit{human–LLM mixed} (contains human-written news and LLM-generated news) dataset.

For the LLM-generated datasets, PolitiFact++ and GossipCop++ are constructed by generating news summaries and headlines using \textit{GPT-3.5}, based on the human-written PolitiFact and GossipCop datasets~\cite{[55]}. To further assess LIFE's ability to detect news generated by other LLMs, we construct three new datasets, GenFake-LLMs, following the same procedure as GossipCop++ \citep{[55]}. Specifically, we use \textit{GPT-Neo}, \textit{GPT-J}, and \textit{LLaMA2-7B} to generate news. For the human-LLM mixed dataset, we use the VLFPN~\cite{[71]}, in which LLM-generated news is generated by \textit{GPT-4} and is highly similar in style to human-written news. The statistics of the LLM-generated datasets are shown in Table \ref{tab:datasets}.

\begin{table}[h]
\centering
\caption{The statistics of fake news datasets.}
\label{tab:dataset_stats}
\begin{tabular}{lccc}
\toprule
\textbf{Dataset} & \textbf{Fake News} & \textbf{Real News} & \textbf{Total}  \\
\midrule
    PolitiFact++ (2024) & 97 & 132 & 229 \\
    GossipCop++ (2024)& 4,084 & 4,169 & 8,253 \\
    VLFPN (2024)& 375 & 400 & 775 \\ \hline
    GenFake-GptNeo & 4,084 & 4,169 & 8,253 \\
    GenFake-GptJ & 4,084 & 4,169 & 8,253 \\
    GenFake-LLaMA2 & 4,084 & 4,169 & 8,253 \\
\bottomrule
\end{tabular}
\label{tab:datasets}
\end{table}

\subsubsection{Baselines}
In this work, we compare ten representative baselines for text-based fake news detection:

\textbf{TextCNN (2014)} \citep{TextCNN} is a model specifically designed for analyzing and processing textual data, utilizing convolutional neural networks. The outputs from all layers are concatenated and fed into a classifier.

\textbf{HAN (2016)} \citep{[26]]} is a Hierarchical Attention Network. In our implementation, we use two GRU layers with 25 hidden units each, followed by two self-attention layers.

\textbf{dEFEND/c (2019)} \citep{[24]]} is a variant of the dEFEND model with the comment-processing module removed. It employs an RNN-based architecture combined with a joint attention mechanism for detection. We set the dimension of the self-attention layer to 100.

\textbf{BERT (2019)} \citep{[53]} is a pre-trained language model. We use it with a task-specific MLP to predict news veracity.

\textbf{RoBERTa (2019)} \citep{[56]} is a pre-trained transformer model enhanced through dynamic masking during training. We use it with a task-specific MLP to predict news veracity.

\textbf{L-Defense (2024)} \citep{[61]} incorporates an unsupervised context learning stage to capture local contextual features, which are then fused with global semantic representations to learn a comprehensive contextual embedding.

\textbf{SheepDog (2024)} \citep{[38]]} enhances robustness against stylistic attacks by generating multi-style news variants via LLMs. It applies consistency regularization to ignore stylistic differences and uses LLM-generated credibility signals for auxiliary supervision.

\textbf{BREAK (2025)} \citep{[60]} introduces a broad-range semantic model by utilizing a fully connected graph to capture holistic semantics, along with dual denoising modules to mitigate noise.

\textbf{ChatGLM2-6B (2023)} \citep{[72]} and \textbf{LLaMA2-7B (2023)} \citep{[57]} are two large language models (LLMs) with 6B and 7B parameters, respectively. We utilize the same training datasets to fine-tune them (using LoRA \citep{[73]}) as the baselines for comparison.

\subsubsection{Implementation Details}

Our method, LIFE, is implemented in PyTorch and trained on an NVIDIA GeForce RTX 4090 GPU. Each dataset is split into training and testing sets with an 80:20 ratio. For baseline models, we adopt the parameter settings as specified in their original publications. For our approach, BERT is employed as the key sentence extraction model with a maximum input length of 512 tokens. It is fine-tuned on the training data with a learning rate of 0.001. During key sentence extraction, we select the top-\(k = 10\) sentences. For probabilistic reconstruction, inference is performed using LLaMA2-7B. In the classification phase, we use a learning rate of \(5 \times 10^{-5}\), weight decay of 0.1, and a warm-up ratio of 0.1.

\subsection{Performance Comparison (RQ1)}
To evaluate the performance of LIFE on LLM-generated fake news detection, and whether it can deal with the real scene of the mixture of human-written news and LLM-generated news. We compare it with ten classical or advanced baselines on the PolitiFact++, GossipCop++, and VLFPN, as shown in Table \ref{tab:performance}.

LIFE achieves the highest performance across all metrics, achieving notable improvements of 1.07\% and 2.9\% in accuracy scores compared to the sub-optimal results on the PolitiFact++ and GossipCop++ datasets, respectively. These results demonstrate the effectiveness of LIFE in capturing prompt-induced linguistic fingerprints for detecting LLM-generated fake news. Additionally, LIFE achieves the best performance on the human-LLM mixed dataset, showing strong practical capability even in real scenarios where news is not all generated by LLMs.

Moreover, CNN-based and RNN-based models (e.g., TextCNN, HAN) have achieved poor results, indicating that traditional methods for human-written fake news datasets may not be well adapted to LLM-generated fake news. For the LLM baselines, we only adopt ChatGLM2-6B and LLaMA2-7B, whose training data are cut off at 2024. We additionally provide results with a stronger recent LLM as a reference in Appendix \ref{RQ:D.1}. LLaMA2-7B exhibits superior detection results compared to ChatGLM2-6B. We attribute this improvement to the fact that LLaMA2-7B is trained on a larger English corpus than ChatGLM2-6B. For BREAK, although it is the latest method, it has not achieved significant results, which may be due to the relatively simple news structure generated by LLM, weakening the advantages of graph denoising. On the contrary, L-Defense and SheepDog, as models for detecting fake news generated by LLM, achieved superior performance, achieving accuracies of 0.879 and 0.908 on VLFPN and GossipCop++ datasets, respectively. However, they often need to introduce a large amount of external knowledge for data augmentation, which incurs inference time costs. Compared to them, our LIFE only requires one original token reconstruction using mLLM in key news fragments to achieve similar or better results.

\begin{table}[t]
\centering
\caption{Performance comparison on real-world datasets.}
\label{tab:realdata}
\small
\begin{tabular}{lcccccc}
\toprule
\multirow{2}{*}{Method} & \multicolumn{2}{c}{PolitiFact++} & \multicolumn{2}{c}{GossipCop++} & \multicolumn{2}{c}{VLFPN}  \\
\cmidrule(lr){2-3} \cmidrule(lr){4-5} \cmidrule(lr){6-7} 
  & Acc. & F1 & Acc. & F1 & Acc. & F1 \\
\midrule
TextCNN    & 0.655 & 0.519 & 0.790 & 0.788 & 0.623 & 0.625  \\
HAN     & 0.633 & 0.491 & 0.793 & 0.778 & 0.589 & 0.578 \\
dEFEND/c & 0.689 & 0.685 & 0.802 & 0.793 & 0.662 & 0.675 \\
BERT        & 0.824 & 0.812 & 0.875 & 0.886 & 0.781 & 0.804  \\
RoBERTa    & 0.841 & 0.830 & 0.897 & \underline{0.908} & 0.764 & 0.821  \\
L-Defense  & 0.881 & \underline{0.882} & 0.865 & 0.864 & \underline{0.879} & 0.870 \\
SheepDog    & \underline{0.883} & 0.876 & \underline{0.908} & 0.894 & 0.867 & \underline{0.889}  \\
BREAK      & 0.860 & 0.859 & 0.900 & 0.901 & 0.873 & 0.865 \\ \hline
ChatGLM2-6B  & 0.836 & 0.824 & 0.885 & 0.879 & 0.792 & 0.808 \\
LLaMA2-7B    & 0.848 & 0.836 & 0.892 & 0.887 & 0.801 & 0.816 \\ \hline
LIFE & \textbf{0.900} & \textbf{0.882} & \textbf{0.937} & \textbf{0.924} & \textbf{0.890} & \textbf{0.926} \\ \hline
Improve (\%)      & +1.700 & +0.000 & +2.900 & +1.610 & +1.100 & +3.700 \\
\bottomrule
\label{tab:performance}
\end{tabular}
\end{table}

\begin{table}[t]
\centering
\caption{Performance comparison on GenFake datasets.}
\label{tab:genfake}
\small
\begin{tabular}{lcccccccc}
\toprule
\multirow{2}{*}{Method} & \multicolumn{2}{c}{GenFake-GptNeo} & \multicolumn{2}{c}{GenFake-GptJ} & \multicolumn{2}{c}{GenFake-LLaMA2} \\
\cmidrule(lr){2-3} \cmidrule(lr){4-5} \cmidrule(lr){6-7}
 & Acc. & F1 & Acc. & F1 & Acc. & F1 \\
\midrule
TextCNN   & 0.801 & 0.796 & 0.708 & 0.706 & 0.764 & 0.752 \\
HAN       & 0.791 & 0.784 & 0.737 & 0.714 & 0.776 & 0.778 \\
dEFEND/c & 0.804 & 0.785 & 0.738 & 0.737 & 0.795 & 0.777 \\
L-Defense & 0.896 & 0.882 & 0.835 & 0.834 & 0.850 & 0.850 \\
SheepDog  & 0.925 & 0.913 & 0.885 & 0.885 & 0.908 & 0.907 \\
BREAK     & 0.880 & 0.879 & 0.850 & 0.849 & 0.879 & 0.879 \\ \hline
LIFE       & \textbf{0.945} & \textbf{0.944} & \textbf{0.916} & \textbf{0.922} & \textbf{0.948} & \textbf{0.945} \\ \hline
Improve (\%)      & +2.000 & +3.100 & +3.100 & +3.700 & +4.000 & +3.800 \\
\bottomrule
\label{tab:generalization}
\end{tabular}
\end{table}
\subsection{Generalization Performance (RQ2)}
The existing open-source datasets are mostly generated by GPT-3.5. In order to test the generalization ability of LIFE in generating news in other series of LLMs, we used GPT-Neo (2.7B), GPT-J (6B), and LLaMA2-7B to generate three datasets according to the approach of Su et al. \citep{[55]}, and the results are shown in Table \ref{tab:generalization}.

LIFE achieves the highest performance across all metrics designed for fake news detection, achieving notable improvements of 2.0\%, 3.1\%, and 4.0\% in accuracy scores compared to the sub-optimal results on the GenFake-GptNeo, GenFake-GptJ and GenFake-LLaMA2 datasets, respectively. These improvements demonstrate that LIFE does not depend on the stylistic patterns of a single LLM but instead captures more generalizable features for distinguishing human- and machine-generated news. This robustness is crucial as generative models evolve rapidly, ensuring that LIFE remains reliable across different architectures. The consistent gains in both accuracy and F1 further indicate balanced improvements in precision and recall, reducing both false positives and false negatives. The results validate LIFE’s strong generalization ability and its potential as a robust solution for future LLM-generated fake news detection scenarios.


\subsection{Ablation Study (RQ3)}
To evaluate the contribution of each component in LIFE, we select four datasets generated by different LLMs for comprehensive analysis. Then, we conduct an ablation study by comparing the full model with four variants: w/o MP, w/o KF, w/o CNN, and w/o TRM. Specifically, w/o MP removes the malicious prompt during reconstruction; w/o KF excludes the key fragment extraction module; w/o CNN removes the CNN branch from the final classifier; and w/o TRM removes the Transformer branch from the final classifier.

\begin{figure}[t]
    \centering
    \includegraphics[width=\columnwidth]{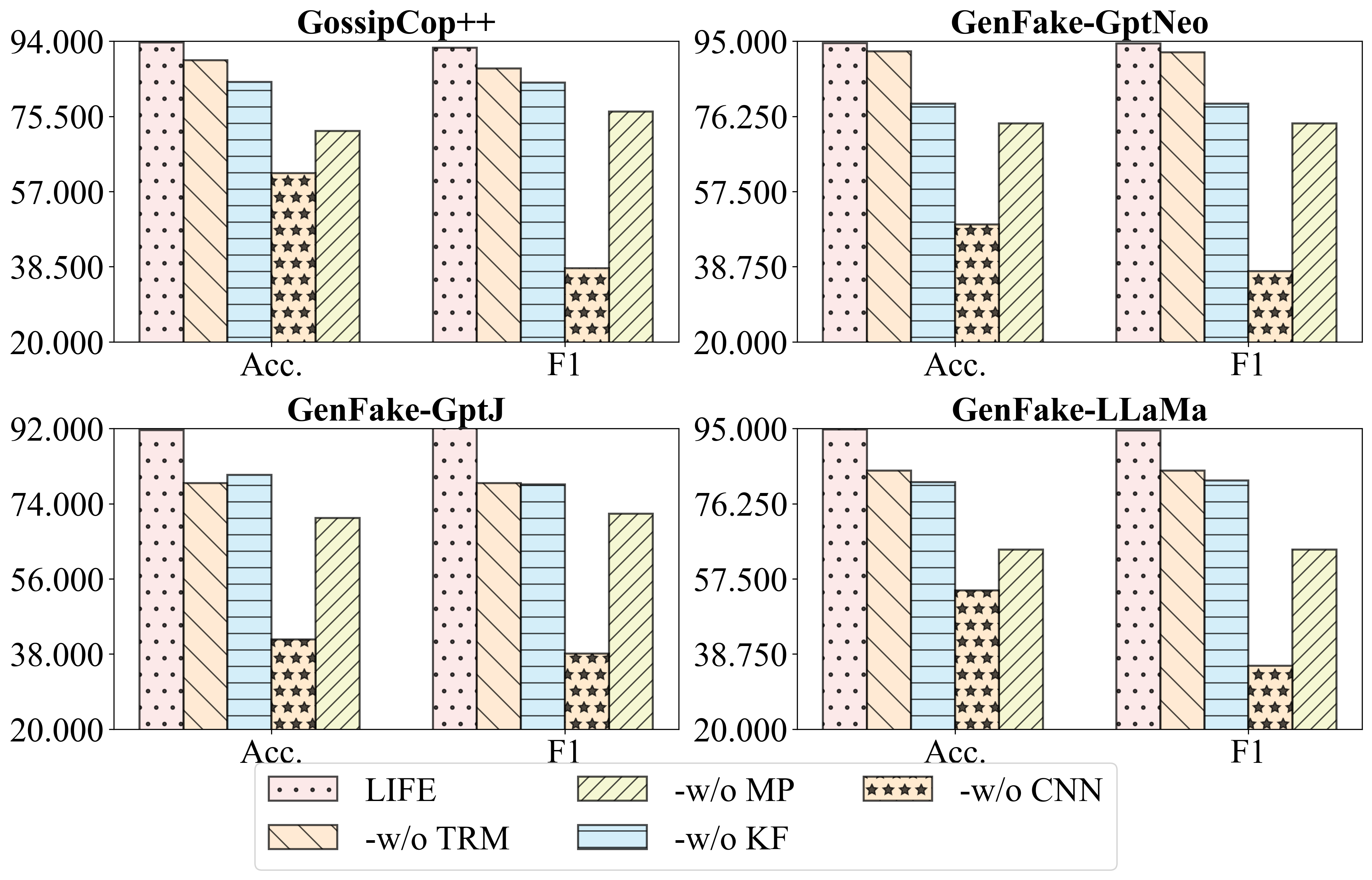} 
    \caption{Ablation study on four datasets.}
    \label{fig:ab} 
\end{figure}

As shown in Figure \ref{fig:ab}, the absence of any part of LIFE leads to sub-optimal performance, indicating that each component contributes positively to the model. In detail, the results of w/o MP show that the LLM reconstructs news based only on semantic information, without fully utilizing the inherent linguistic fingerprints, resulting in poor performance in the news detection task. Moreover, the performance of w/o KF is better than w/o MP, indicating that the guidance provided by the malicious prompt to the LLM is more important than merely removing noise from the text (Additional experiments on the key-fragment extractor are provided in Appendix \ref{RQ:D.2}). These experimental findings demonstrate the effectiveness of both malicious prompt guidance and textual denoising.

Meanwhile, both w/o CNN and w/o TRM result in performance degradation, with w/o CNN performing the worst. The particularly poor performance of w/o CNN indicates that, due to the input features being one-dimensional, the Transformer layers fail to effectively learn useful classification patterns. In contrast, w/o TRM achieves relatively better performance, showing that the CNN structure is effective in handling probabilistic vector features.

\subsection{Case Study (RQ4)}
\begin{figure*}[t]

    \centering
    \subfigure[The visualization of fake news sentences’ weights.]{
        \includegraphics[width=0.48\textwidth]{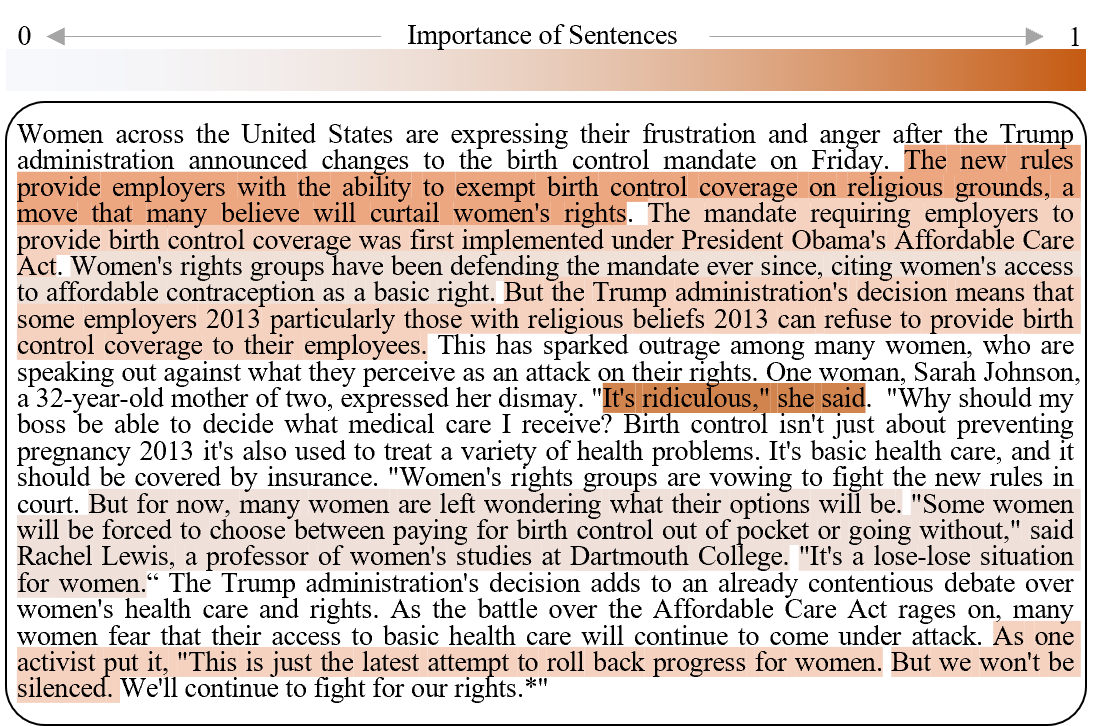}
    }
    \hfill
    \subfigure[The visualization of real news sentences’ weights.]{
        \includegraphics[width=0.48\textwidth]{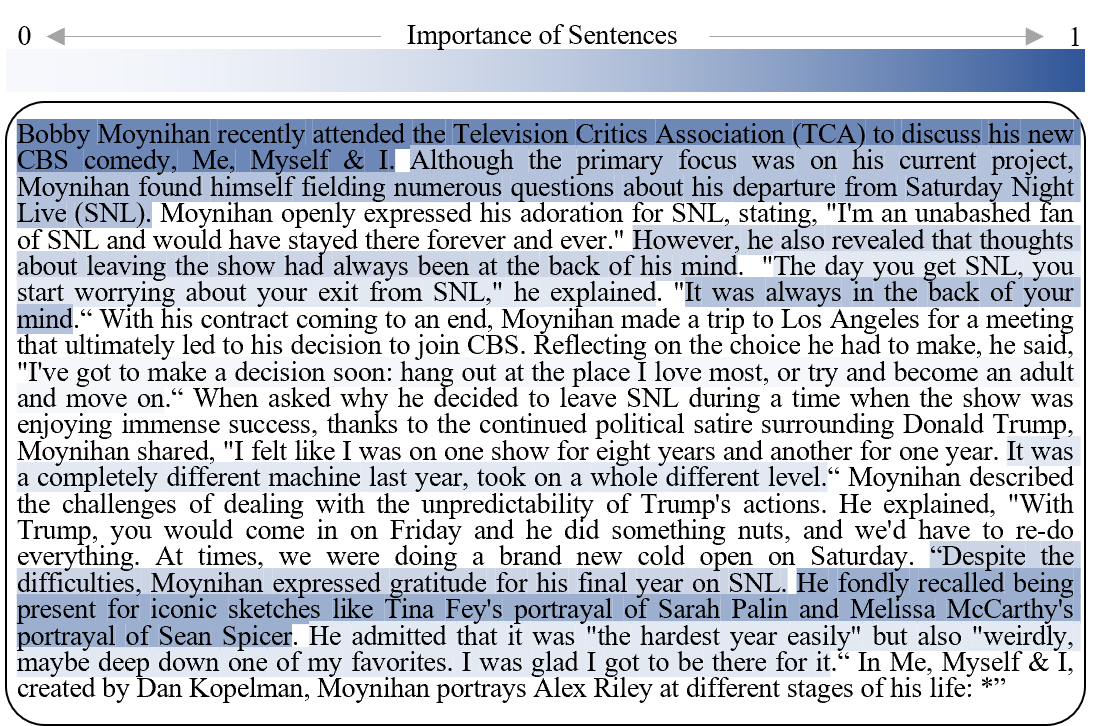}
    }

    \subfigure[The probabilities of all words in real and fake news.]{
        \includegraphics[width=0.48\textwidth]{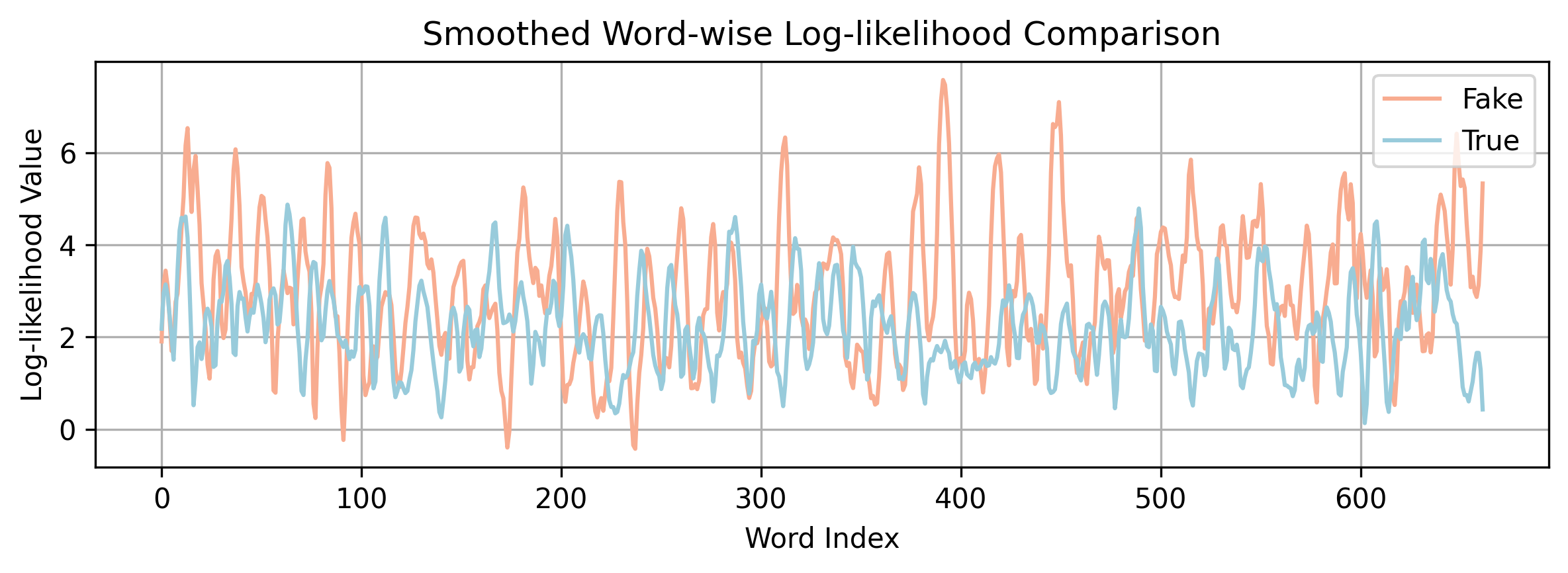}
    }
    \hfill
    \subfigure[The probabilities of key fragment words in real and fake news.]{
        \includegraphics[width=0.48\textwidth]{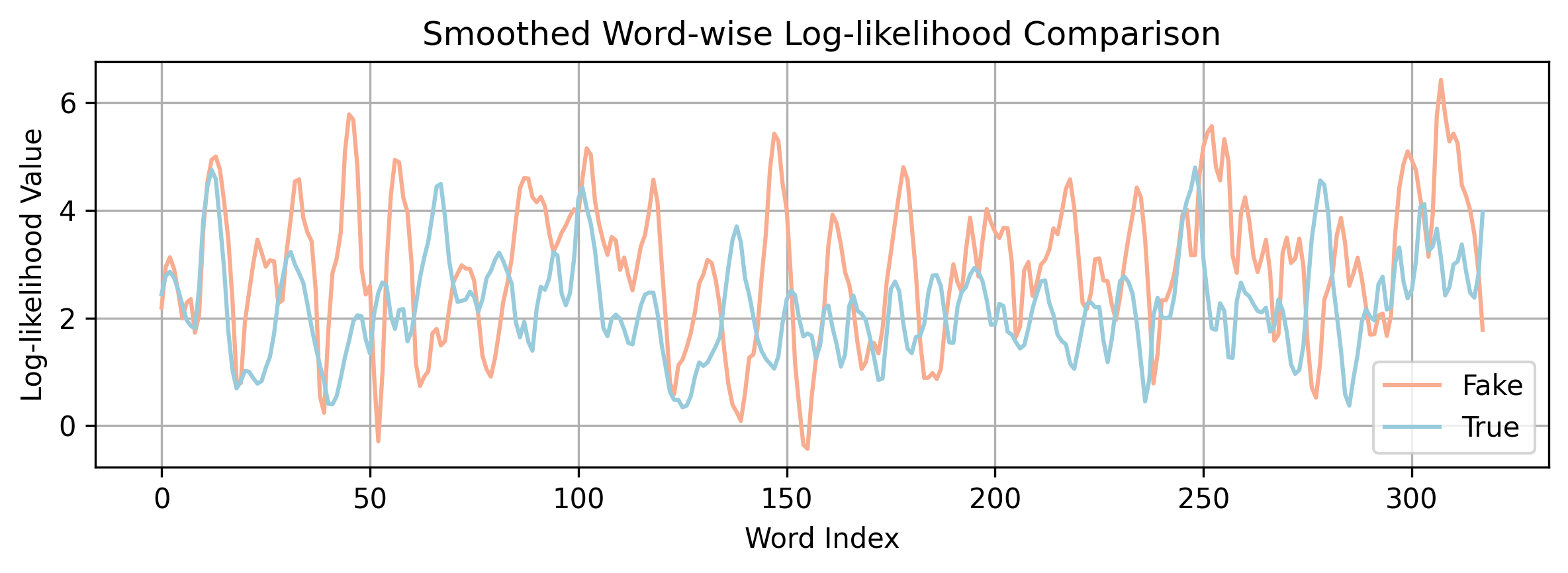}
    }
    
    \caption{A case study on the differences in key sentences and reconstruction probability distributions. (a) and (b) show the visualizations of key sentences from real and fake news, respectively. (c) and (d) compare the reconstruction probabilities of all words in the complete news articles and the corresponding key sentences.}
    \label{fig:case}
\end{figure*}

To further illustrate how LIFE distinguishes between real and fake news, we analyze the reconstruction probability distributions of both the full text and key fragments, as depicted in Figure~\ref{fig:case}.

We first obtain the word-level reconstruction probabilities of an entire real and fake news article using the mLLM, and plot these values as sequential feature curves, where the x-axis represents word positions and the y-axis denotes the corresponding reconstruction probabilities (see Figure~\ref{fig:case}(c)). The results show that in the segments from position 0 to 100 and 400 to 600, fake news exhibits higher reconstruction probabilities than real news. However, between positions 100 and 400, the two curves frequently overlap, indicating the presence of redundant content. This aligns with typical news-writing patterns, where critical information tends to be concentrated at the beginning and end of an article.

Correspondingly, Figure~\ref{fig:case}(a) and Figure~\ref{fig:case}(b) show that LIFE tends to select sentences from the beginning and end as key fragments, suggesting that the model may implicitly rely on reconstruction probability variation when identifying semantically important regions. Figure~\ref{fig:case}(d) illustrates that the reconstruction probability differences between real and fake news are more distinct and stable when focused on key fragments. This further confirms the effectiveness of key fragment selection in enhancing the discriminative features used by LIFE.

\section{Related Work}
\subsection{Fake News Detection}
The task of fake news detection involves identifying news that contains false information with potential harm \citep{[39]]}. Traditional methods rely on expert consultation for manual detection, which is both expensive and time-consuming \citep{[41],[42]}.

To reduce cost, researchers automate detection using machine learning and deep learning techniques. Early studies exploited hierarchical text structure and high-frequency word features \citep{[24]],[25]]}. Later work improved feature extraction by incorporating sentiment word frequency, positional features, and auxiliary signals from texts \citep{[30]],[28],[29]}. Recent studies emphasize robustness alongside accuracy, since real-world supervision can be noisy, biased, or adversarially manipulated \citep{[78],[77]}. Xu et al. \citep{[82]} estimate label-specific noise patterns and perform fine-grained text corrections.

With the rise of LLMs, researchers leverage them for fake news detection \citep{[39]],[44]}. Early zero-shot querying for labels yields limited success \citep{[18]],[74]}. More recent efforts use LLMs to generate auxiliary information while relying on smaller fine-tuned models for classification \citep{[45],[46]}. However, these pipelines still depend on static feature extraction, and may struggle as LLM-generated news becomes more fluent and diverse, particularly when generators adapt their style and coherence to appear increasingly human-written.

\subsection{LLM-Generated Fake News Detection}
With the continuous advancement of LLMs, an increasing amount of news content is being produced using LLMs \citep{[75],[6]]}. This trend can harm recommendation systems and other real-world applications \citep{[79],[80],[81]} by amplifying the spread of misinformation, undermining public trust, and distorting user engagement signals.

Initially, most LLM-generated news articles were created by malicious users, which led some researchers to adopt AI text detection as a direct approach to identify LLM-generated fake news \citep{[15]}. However, as LLMs become more widely used in news creation, solely detecting LLM-generated text has become inadequate. Existing methods designed for detecting human-written fake news show a significant decline in performance when applied to LLM-generated fake news detection \citep{[37]],[18]]}. To address the challenge of detecting LLM-generated news, Shu et al. \citep{[44]} analyzed various types and challenges of LLM-generated misinformation. Wu et al. \citep{[38]]} improved robustness against disinformation generated by LLMs by learning from a variety of attack styles produced by LLMs. Some other works have proposed methods addressing the diversity of LLM-generated news, focusing on continuous prompts \citep{[47],[48]}. However, the aforementioned methods rely heavily on article content for detection, which makes them sensitive to domain shifts and necessitates frequent prompt optimization \citep{hong2025beyond}. Therefore, we shift the focus to the internal generation mechanisms of LLMs themselves.

\section{Conclusion}
In this paper, we propose LIFE, a probabilistic framework for detecting LLM-generated fake news by exploiting differences in generation behavior. We identify a consistent prompt-induced linguistic fingerprint, where malicious prompts lead LLMs to assign higher token-level probabilities to fake news. LIFE models this phenomenon through probabilistic reconstruction, key fragment extraction, and a CNN--Transformer classifier. Experiments on multiple LLM-generated and human-LLM mixed datasets show that LIFE outperforms existing detectors and maintains strong cross-model generalization. The results demonstrate that analyzing the generative process provides a more reliable foundation for fake news detection than relying solely on surface linguistic features.


\begin{acks}
This work is supported by the National Natural Science Foundation of China (Grant No. 62176028), the Natural Science Foundation of Chongqing, China (Grant No. CSTB2024NSCQ-MSX0617), Project to Attract Foreign Experts (Grant No. H20251055), and the Royal Society International Exchange Grant (IEC\textbackslash
NSFC\textbackslash
242347) through Royal Society and NSFC.
\end{acks}

\bibliographystyle{ACM-Reference-Format}
\bibliography{sample-base}

\appendix

\section{Information-Theoretic Justification}
\label{Appendix:A}
We formally justify why the \emph{log-odds drop} $\Delta\ell_j$ used in our masking scheme is a valid unsupervised measure of sentence ``keyness''.  
Let a news article be $x=\{s_1,\dots,s_n\}$ with veracity label $y\!\in\!\{0,1\}$.  
Masking sentence $s_j$ produces $x^{(-j)}$, and we define
\begin{equation}
\Delta\ell_j = \log\frac{p(y{=}1\mid x)}{p(y{=}0\mid x)} - \log\frac{p(y{=}1\mid x^{(-j)})}{p(y{=}0\mid x^{(-j)})}.
\label{eq:logodds}
\end{equation}

By the definition of conditional mutual information, removing $s_j$ induces an information loss about $Y$:
\begin{equation}
I(s_j;Y\mid x\setminus s_j)
=\mathbb{E}_{x,y}\!\left[\mathrm{KL}\!\big(p(y\mid x)\,\|\,p(y\mid x^{(-j)})\big)\right],
\label{eq:cmi}
\end{equation}
where $\mathrm{KL}(\cdot\|\cdot)$ is the Kullback–Leibler divergence.  
For a specific sample, define the instance-level loss
\begin{equation}
\Delta\mathcal{I}_j=\mathrm{KL}\!\big(p(y\mid x)\,\|\,p(y\mid x^{(-j)})\big).
\end{equation}
In the binary case $p(y{=}1\mid x)=\sigma(\ell(x))$, with $\sigma(\cdot)$ the logistic function and natural parameter $\ell(x)$ (the posterior log-odds).  
If masking $s_j$ changes the log-odds by $\Delta\ell_j$, i.e. $\ell(x^{(-j)})=\ell(x)-\Delta\ell_j$, the Bernoulli KL divergence has closed form:
\begin{equation}
\Delta\mathcal{I}_j
= p\log\frac{p}{q} + (1-p)\log\frac{1-p}{1-q},\qquad
p=\sigma(\ell),\ q=\sigma(\ell-\Delta\ell_j).
\label{eq:bernkl}
\end{equation}
A second-order Taylor expansion of~\eqref{eq:bernkl} around $q=p$ gives
\begin{equation}
\Delta\mathcal{I}_j \approx \tfrac{1}{2}\,p(1-p)\,(\Delta\ell_j)^2,
\label{eq:quad}
\end{equation}
where $p(1-p)$ is the Fisher information of the Bernoulli distribution.  
Equation~\eqref{eq:quad} implies that $\Delta\mathcal{I}_j$ is a monotonic function of $|\Delta\ell_j|$, hence maximizing $|\Delta\ell_j|$ equivalently maximizes the conditional mutual information $I(s_j;Y\mid x\setminus s_j)$ in expectation.  
Therefore, ranking sentences by $|\Delta\ell_j|$ selects those whose removal causes the largest expected information loss about the veracity label.

\noindent\textbf{Interpretation.}  
$\Delta\ell_j$ measures how much the posterior decision boundary would shift if $s_j$ were absent; it is thus a natural, label-free measure of \textit{decision informativeness}.  
Using log-odds rather than raw probability drops ensures numerical stability and aligns with the natural parameterization of the Bernoulli posterior.

\noindent\textbf{Implication for LLM reconstruction.}  
Sentences with large $|\Delta\ell_j|$ carry concentrated semantic and factual content that most strongly influences veracity prediction.  
When these key fragments are fed to an LLM for prompt-induced reconstruction, they correspond to regions of high conditional likelihood curvature areas where generative models exhibit the greatest difference between genuine and fabricated news.  
Consequently, BERT-derived key fragments amplify the reconstruction-probability gap exploited in our detector, enhancing detection accuracy while reducing redundancy.

\begin{figure}[!t]
    \centering
    \includegraphics[width=\columnwidth]{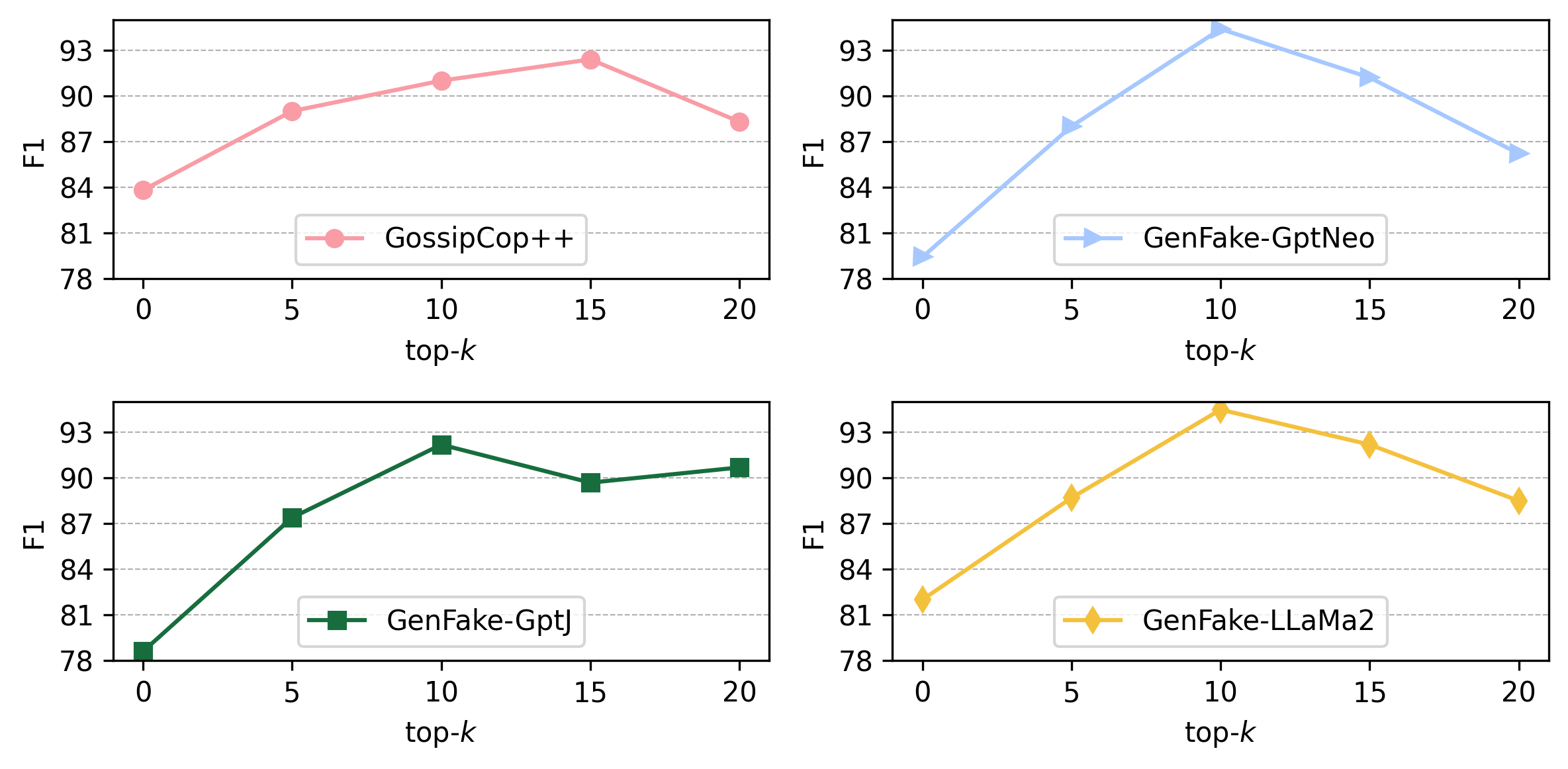} 
    \caption{Hyperparameters sensitivity with regard to \(k\).}
    \label{fig:hyper} 
\end{figure}
\section{Sensitivity of Hyperparameter k (RQ5)}
\label{RQ5}
In the \textsc{LIFE} model, the hyperparameter $k$ determines the number of top-ranked key fragments retained during the reconstruction process. Its primary role is to filter out redundant information and amplify reconstruction differences between real and fake news. Consistent with RQ2, we select four datasets generated by different LLMs to analyze the influence of $k$ across diverse scenarios.

We analyze the impact of varying \(k\) and visualise the results in Figure~\ref{fig:hyper}. Here, \(k = 0\) denotes the absence of key fragment selection. The results show that LIFE achieves the best performance on the GossipCop++ dataset when \(k = 15\), and attains optimal results on the other three datasets when \(k = 10\). This demonstrates the effectiveness of selecting top-\(k\) fragments, as well as the necessity of focusing on semantically important regions.

Moreover, we observe a clear performance improvement as \(k\) increases from 0 to 20, indicating the value of key fragment selection in enhancing detection. However, setting \(k\) too high introduces redundant information, which weakens the discriminative power of the model. These findings highlight the importance of carefully tuning \(k\) to balance information preservation and noise reduction.

\section{The Impact of Malicious Templates (RQ6)}
\label{RQ6}
\subsection{Definition and Design of Malicious Prompts}
We formally define a \textit{malicious prompt} as a conditional perturbation function that intentionally guides a large language model (LLM) to reconstruct factual news into misinformation-oriented text. 
Given an original news piece $x$, the malicious prompt $p_m$ operates as:
\begin{equation}
x' = \mathrm{LLM}(p_m \oplus x),
\end{equation}
where $\oplus$ denotes prompt concatenation and $x'$ represents the reconstructed (potentially fake) news text. 
The prompt $p_m$ explicitly induces deviations in factual correctness, sentiment polarity, or source credibility.

\paragraph{Functional Categories.}
To systematically capture manipulation intents, we categorize malicious prompts into three functional types:
\begin{itemize}
    \item \textbf{Factual Distortion (FD):} alters or fabricates factual claims, e.g., ``rewrite this news to distort the main fact.'' 
    \item \textbf{Emotional Amplification (EA):} exaggerates emotional or ideological polarity, e.g., ``make the article sound more shocking or partisan.'' 
    \item \textbf{Contextual Fabrication (CF):} introduces fabricated context, sources, or expert opinions, e.g., ``add expert comments supporting the claim.'' 
\end{itemize}

Each dimension corresponds to one manipulation axis in the latent news space, ensuring coverage of the major pathways of fake news generation. 

\paragraph{Prompt Template Schema.}
We design a broader pool of malicious prompt templates following a compositional schema
$T_i = [\text{Action}] + [\text{Intent(dim, strength)}] + [\text{Constraint}]$,
covering three manipulation dimensions (FD/EA/CF). 
In this paper we quantitatively evaluate three representative templates (T1--T3) to compute the budget and to avoid overfitting to prompt surface forms.

\paragraph{Representative Examples.}
We list three representative templates from the constructed pool below:

\begin{promptbox}
T1 (Factual Distortion): Rewrite the following article to make it appear completely authentic and confident fake news. Avoid uncertain expressions or ambiguous language.
\end{promptbox}

\begin{promptbox}
T2 (Emotional Amplification): Reconstruct fake news with a strong emotional tone to make readers feel convinced or angry. Maintain sentence fluency and confidence.
\end{promptbox}

\begin{promptbox}
T3 (Contextual Fabrication): Using authoritative statements or fictional expert opinions to expand into fake news, making the story more credible. Maintain a professional news style.
\end{promptbox}

\begin{table}[!t]
\caption{Prompt performance comparison on the GossipCop++ dataset. Bolded values indicate the best results.}
\centering
\small
\begin{tabular}{lcccc}
    \toprule
    \textbf{Methods} & \textbf{Accuracy} & \textbf{Precision} & \textbf{Recall} & \textbf{F1} \\ 
    \midrule
    LIFE (T1) & 0.929 & \textbf{0.925} & 0.906 & 0.915 \\
    LIFE (T2) & \textbf{0.940} & 0.912 & \textbf{0.931} & \textbf{0.921} \\
    LIFE (T3) & 0.934 & 0.923 & 0.909 & 0.917 \\
    \midrule
    \textbf{Max difference} & \textbf{1.100\%} & \textbf{1.300\%} & \textbf{2.200\%} & \textbf{0.600\%} \\
    \bottomrule
\end{tabular}
\label{tab:5}
\end{table}

\subsection{Result Analysis and Robustness}
Table~\ref{tab:5} reports LIFE's performance under three representative malicious prompts (T1--T3) on GossipCop++. 
We observe a small variation across prompts (max accuracy difference $=1.1\%$), indicating that prompt-induced linguistic fingerprints are stable across distinct manipulation intents.
To further quantify stability, we compute the cosine similarity between the reconstructed probability vectors and obtain consistently high values across T1--T3.
A benign paraphrasing control (no disinformation intent) yields a higher Jensen--Shannon divergence against malicious reconstructions, supporting that malicious intent, rather than mere rewording, drives the detectable probability-shift patterns.

\subsection{Additional Prompt Robustness Analyses}
To further strengthen the robustness evaluation with respect to prompt design, beyond the three representative malicious-prompt templates summarized in C.1, we additionally adopt the six intentional misinformation-generation prompt templates summarized by \citet{[14]]}. Specifically, these templates include Totally Arbitrary Generation, Partially Arbitrary Generation, Paraphrase Generation, Rewriting Generation, Open-ended Generation, and Information Manipulation. Across six prompt types, we observe a consistent probability gap between fake and real news, and the resulting detection performance is stable, with F1 varying within $\pm 1.00\%$ on GossipCop++.

To probe robustness to counter-prompt, we append a representative style-mimicking suffix (i.e., ``write in the style of real news'') to the malicious prompt, while keeping the rest of the pipeline unchanged. In this setting, LIFE still outperforms the strongest baseline in our experiments (F1(LIFE) = 0.901 > F1(SheepDog) = 0.885). This result provides preliminary evidence that the probability-gap signal remains useful under this type of counter-prompt suffix.

\section{Additional Analyses}
In response to concerns about (i) whether our LLM-based baselines are sufficiently strong and up-to-date, and (ii) whether our performance depends on the specific key-fragment extractor, we conduct two additional experiments. First, we include a stronger recent open-source LLM (Qwen3-8B \citep{yang2025qwen3}) as an additional reference baseline. Second, we replace the key-fragment extraction module with a simple LLM-based alternative to assess the sensitivity of the overall pipeline to the extractor choice.

\subsection{A stronger recent LLM as a reference}
\label{RQ:D.1}
The main paper restricts LLM-based detection baselines to models whose pre-training cutoff is less likely to overlap with our 2024 datasets, thereby reducing potential fairness and data-leakage issues. To provide a stronger contemporary reference, we additionally evaluate Qwen3-8B, a recent open-source LLM. Given that its pre-training data may extend into 2024, we treat this result as a reference comparison rather than a strictly fair supervised detection baseline. On VLFPN dataset, the LLM baselines achieve F1 scores of 0.808 for ChatGLM2-6B and 0.816 for LLaMA2-7B, while the stronger recent model Qwen3-8B attains 0.872. The improvement of Qwen3-8B over the two earlier baselines may be attributed to its more recent training data (released in 2025) as well as stronger model design and scaling. Nevertheless, Qwen3-8B remains below LIFE, which achieves an F1 score of 0.926.

\subsection{Sensitivity to the key-fragment extractor}
\label{RQ:D.2}
To evaluate the sensitivity of the key-fragment extraction module, we compare our original extractor with a simple LLM-based alternative that prompts the LLM to select the ten sentences it deems most important for fake-news detection for each article in VLFPN. The resulting key-fragment sets partially overlap, with an average overlap ratio of 0.732. We define the overlap ratio as the fraction of sentences selected by both extractors, averaged over articles. The overall detection performance changes only modestly, with F1 decreasing from 0.900 to 0.881, suggesting limited sensitivity to the extractor choice in this tested setting.

\end{document}